\documentclass[11pt]{article}

\usepackage[utf8]{inputenc}
\usepackage[T1]{fontenc}
\usepackage{lmodern}
\usepackage{microtype}
\usepackage[letterpaper,margin=1in]{geometry}
\usepackage{amsmath,amssymb}
\usepackage{graphicx}
\usepackage{booktabs}
\usepackage{array}
\usepackage{tabularx}
\usepackage{enumitem}
\usepackage{caption}
\usepackage{float}
\usepackage{titlesec}
\usepackage{xcolor}
\usepackage[colorlinks=true, linkcolor=blue!50!black, citecolor=blue!50!black, urlcolor=blue!50!black]{hyperref}

\titlespacing*{\section}{0pt}{1.4ex plus 0.4ex minus 0.2ex}{0.8ex plus 0.2ex}
\titlespacing*{\subsection}{0pt}{1.0ex plus 0.3ex minus 0.2ex}{0.4ex plus 0.1ex}
\titlespacing*{\subsubsection}{0pt}{0.8ex plus 0.2ex minus 0.1ex}{0.3ex plus 0.1ex}

\setlist{itemsep=0.2ex, topsep=0.4ex}

\newcommand{\zsig}[1]{$z = #1\sigma$}

\title{Closure-Validated Circuit Discovery in Attention Heads:\\
Co-activation Proposes, Ablation Disposes}

\author{Yongzhong Xu\thanks{Code, full results, and a self-contained writeup:
\url{https://github.com/skydancerosel/coactivation-closure}. Correspondence:
\texttt{abbyxu@gmail.com}.}}
\date{May 2026}

\begin{document}

\maketitle

\begin{abstract}
Interpretability increasingly treats \emph{groups} of components, not
individual units, as the basic object, and proposes to find them by clustering
co-activation statistics. We ask whether such a cheap signal actually
identifies an \emph{attention-head} circuit. Adapting the binarized-Ising
clustering recipe of Bhalla et al.~\cite{bhalla2026} to attention heads ---
but validating by causal ablation rather than by the reconstruction criterion
used for sparse-autoencoder feature manifolds --- we cluster heads and then run
a closure test: ablate the discovered community and compare per-example damage
to matched-random controls on cross-entropy loss, accuracy, and target-token
logit. Across two dense 1B-scale models (Pythia~1B, OLMo~1B) and two input
distributions, the communities pass closure. In a Mixture-of-Experts model
(OLMoE-1B-7B), route-conditional clustering recovers a statistically real
signal ($+3.05\sigma$ above a random-partition null) that nonetheless does not
survive closure --- ablation \emph{improves} loss, in the direction opposite
to a real circuit. Extending closure across training checkpoints, two further
proxies --- attention-target selectivity and participation ratio --- decouple
from function in both directions: function is load-bearing before the attention
pattern forms, and a sharp attention pattern with high participation ratio can
carry no function. We conclude that a cheap signal --- cluster membership,
attention selectivity, or participation ratio --- is a circuit \emph{proposal},
not a confirmed circuit; closure is what separates the two, and target-token
logit and accuracy read closure more reliably than aggregate loss.
\end{abstract}

\section{Introduction}

Mechanistic interpretability has converged, in recent years, on an
ensemble-of-units view of how neural networks compute. Sparse autoencoders
find that concepts are tiled across many features rather than localized to
one~\cite{bhalla2026}; attention behaviors are distributed across multiple
heads rather than implemented by a canonical single
head~\cite{goodfire2026}; continuous concepts (color, day-of-week, age)
traverse low-dimensional manifolds in mid-network
activations~\cite{engels2024,park2024geometry}. Across these results, the
interpretive unit is a group --- a subspace, a manifold, a head-set --- and
not a direction.

A natural methodological move follows: if the interpretive unit is a group,
find groups by clustering. Co-activation statistics across a fixed batch of
inputs become the data; affinity between two units (feature similarity,
mutual information of binary activations, or conditional couplings in a
pairwise Ising model) becomes the proposed edge weight; spectral clustering
or modularity optimization on the resulting affinity graph yields candidate
groups. Bhalla et al.~\cite{bhalla2026} formalize a version of this for SAE
features and recover manifold-tiled groups whose joint activity tracks the
underlying geometry.

The question this paper asks is not whether clustering finds \emph{something}
--- it clearly does --- but whether what it finds is a \textbf{circuit} in
the load-bearing sense of mechanistic interpretability: a group of components
whose removal degrades the model's behavior on inputs where the group's
putative function applies. The distinction matters because the same
clustering output is consistent with several different underlying realities:

\begin{enumerate}
\item A group of heads that jointly implement a function, whose removal
  substantially damages the function on relevant inputs (a \emph{circuit});
\item A group of heads that fire together because they share an upstream
  driver (a shared input, a routing pattern, a positional bias) but
  contribute redundantly or marginally to the output (a \emph{co-active
  community} that is not a circuit);
\item A group of heads that share a co-activation pattern because they all
  add a similar bias signal that the rest of the network compensates for at
  the output (a \emph{cancellable community} whose removal might even help).
\end{enumerate}

The clustering output alone cannot distinguish these. A test that \emph{can}
distinguish them is straightforward: ablate the discovered community by
zeroing the per-head outputs that contribute to the residual stream, compare
per-example damage on the same batch to that produced by matched random-head
ablations, and ask whether the candidate's damage is (a) in the direction
expected of a real circuit (loss up, accuracy down, target-token logit down)
and (b) outside the distribution of effects produced by random head-sets of
the same size.

We run this contrast on three 1B-scale language models --- Pythia~1B,
OLMo~1B, OLMoE-1B-7B --- and two input distributions --- a synthetic induction
batch and a natural-text batch from prior probe-circuit work~\cite{xu2026}.
The dense models, on both distributions, produce co-activation communities
that pass closure: the candidate's ablation damage is in the predicted
direction and significantly above the random-control distribution. The MoE
model, on natural text, does not. Its marginal Ising signal collapses
entirely. Route-conditional stratification --- clustering examples by their
routing pattern and fitting separate Isings within each route stratum ---
recovers a statistical signal that survives a careful random-partition null
control, but the recovered community then fails closure in the wrong
direction: ablation improves loss, more so on inputs outside the stratum the
community was discovered on.

We treat this asymmetry as the paper's main finding. The dense closures show
what clustering-plus-closure can do; the MoE closure shows what
clustering-plus-route-conditioning-plus-closure cannot do, in a case where
the clustering looks methodologically careful and the statistical signal
looks real. The gap between proposal and discovery is architecturally
specific, and it widens to wrong-direction in MoE on natural inputs even
after the obvious fix.

We then ask the same question of two other cheap proxies, across the
training axis. Using cached intermediate checkpoints, we measure when
attention-to-target selectivity and participation ratio emerge for each
head class and when the class becomes load-bearing under closure. The two
do not coincide: BOS-head function is load-bearing $\approx 2$B tokens
before its attention pattern forms in two dense architectures (function
without form), while previous-token heads at one checkpoint show a sharp
attention pattern and high participation ratio with no closure signal
(form without function). The bidirectional crossing shows attention pattern
and function are distinct constructs, not the same construct at different
sensitivities --- the same lesson as the co-activation result, now for two
more proxies.

We also document a multi-metric reading of closure that we recommend as a
standard. Cross-entropy loss, taken alone, would misrank our community-closure
tests: it inflates one positive result by control-variance collapse (OLMo~1B
natural, \zsig{+36}), understates another by metric-aggregation slack
(Pythia~1B natural, \zsig{+1.95} on loss, \zsig{-52.5} on target-logit), and
gives a numerically large but qualitatively wrong-direction value for the MoE
failure. Target-token logit and top-1 accuracy each give a more stable
ranking and combine to give a coherent verdict. The Pythia~1B natural-text
divergence between metrics is itself an architectural finding: a $25\times$
gap in $z$-scores within the same test on the same ablation. We read it as a
downstream-redundancy signature --- the discovered circuit is real and
specific, but the model has redundant downstream pathways that partially
reconstruct the ablated community's output, so the loss aggregate is
partially insulated even while the per-target logit collapses.

The paper exhibits five community-closure tests as its primary evidence
base (\S\ref{sec:dense}--\S\ref{sec:moe}, summarized in
\S\ref{sec:verdict}), extends the closure test across training to two
further proxies (\S\ref{sec:developmental}), and collects its claims and an
explicit \emph{what we do not claim} scope statement in
\S\ref{sec:discussion} --- the temptation to overgeneralize from this
evidence is real and we mark out the scope carefully.

\section{Related work}

\paragraph{Co-activation and group-level interpretability.}
Bhalla et al.~\cite{bhalla2026} develop a framework in which
sparse-autoencoder features are grouped post-hoc by conditional
co-activation, fitting a pairwise Ising model on binarized SAE codes and
recovering feature communities that align with continuous concept manifolds
(age, color, day-of-week, temperature). Their pipeline is the most direct
methodological analog to what we test here, and we adapt their
binarization-plus-Ising recipe to attention-head focus statistics. Two
important differences from their setting: (i) we apply the recipe to a
discrete-circuit domain (attention heads with classifiable capability
functions --- induction, previous-token, first-token, etc.) rather than
continuous concept manifolds; (ii) we ablate the discovered communities to
test their causal load-bearing-ness, whereas Bhalla et al.\ validate
communities by reconstruction coherence (whether atoms grouped together
collectively reconstruct the manifold).

\paragraph{Distributed attention behaviors.}
Goodfire's parameter-decomposition work~\cite{goodfire2026} decomposes
weight matrices into rank-1 subcomponents and observes that single attention
behaviors (previous-token, syntax-boundary) distribute across multiple heads
rather than localizing. This is the parameter-side counterpart to what
clustering observes on the activation side: real attention functions inhabit
subspaces, not single heads.

\paragraph{Linear vs.\ nonlinear representations.}
The Linear Representation Hypothesis~\cite{park2024geometry,park2023lrh}
treats concepts as one-dimensional directions in activation space;
subsequent work~\cite{engels2024,kantamneni2025} finds that this is
insufficient for many concepts (circular topologies for periodic concepts;
polar coordinates for syntactic structure). Our setting is discrete-concept
(attention head classification by capability), so LRH-vs-manifold isn't
directly the issue --- but the underlying philosophical move, that
interpretation lives in subspaces rather than single directions, is shared.

\paragraph{Circuit discovery and validation.}
The mechanistic-interpretability literature on circuits typically combines
(i) a hypothesis about a specific function (induction, IOI, indirect-object
identification), (ii) a probe or attribution graph that identifies candidate
components, and (iii) ablation or path-patching to validate the components'
causal role~\cite{olsson2022, wang2023ioi, conmy2023}. The novel methodological
question we test is whether step (ii) can be replaced --- partially or
wholly --- by \emph{unsupervised} co-activation clustering with no function
specified in advance, with step (iii) serving as the validation gate. The
dense results suggest this is feasible for some functions in dense models;
the MoE result shows where the substitution breaks down.

\paragraph{MoE interpretability.}
Mixture-of-Experts models have been studied primarily through the experts
themselves --- what each expert specializes in, how routing patterns
correlate with input type~\cite{zoph2022}. The attention-head circuits of
MoE transformers are much less studied, and our finding that
route-conditional clustering recovers a statistical signal that fails
closure is, to our knowledge, the first systematic comparison of
co-activation interpretability between dense and MoE architectures.

\section{Method}
\label{sec:method}

\subsection{Overall design}
The study tests four claims (made precise in \S\ref{sec:discussion}) by
running five closure tests. Each closure test consists of three stages:
(i)~\emph{proposal} --- given a model and a batch, cluster attention heads
by co-activation statistics and select the cleanest candidate community;
(ii)~\emph{prediction} --- declare the candidate a circuit if ablation
damages the model's behavior in a predicted direction; (iii)~\emph{closure}
--- ablate and compare per-example damage to five matched random-head
controls, reporting multi-metric $z$-scores. The four dense $\times$
two-distribution tests cover \emph{whether the dense pipeline finds
circuits across model and input distribution}. The fifth (MoE $\times$
natural-text $\times$ route-conditional) covers \emph{whether the obvious
extension to MoE recovers circuits}.

\subsection{Forward extraction}
For each model and input batch, we run a forward pass with
\texttt{output\_attentions=True} and extract the per-(example, layer, head)
attention pattern at a per-example \emph{query position}. The query
position is the last token for the synthetic batch and a per-example index
(saved with the batch from prior work) for the natural batch. The extracted
tensor has shape $(N, L, H, T)$ where $N=2000$ examples, $L$ is the number
of layers, $H$ is the number of heads per layer, and $T=256$ is the
sequence length.

For the MoE forward pass we additionally set
\texttt{output\_router\_logits=True} and extract the softmax routing
weights at the query position per layer, giving a tensor of shape
$(N, L, E)$ where $E=64$ is the number of experts per layer (OLMoE).

\subsection{Template-free signal}
For each example $i$ and head $(L, H)$, we collapse the attention pattern
at the query position to a single scalar:
\[
\text{sig}_{i, (L, H)} \;=\; \max_k\, \text{attn}_{i, L, H, k}.
\]
This is the maximum attention weight assigned by the head to any key
position in this input. High values indicate that the head is \emph{focused}
on something (some particular key); low values indicate a near-uniform
distribution over keys. The signal is \textbf{template-free} in the sense
that no specific key position (BOS, previous-token, induction target, etc.)
is privileged --- only the head's degree of focus matters.

This choice mirrors Bhalla et al.'s use of SAE code magnitudes as their
template-free signal. Alternatives we considered and did not use: attention
entropy (low entropy = focused; equivalent up to monotone transform to max
for most heads), attention to a specific canonical target (e.g., BOS ---
but this hard-codes a hypothesis), and the L2 norm of the head's OV output
(function-aligned but requires extra forward machinery to extract).
Max-attention is the simplest fully template-free signal that uses only the
attention pattern.

\subsection{Per-head median binarization}
For each head $(L, H)$, we binarize its signal across the $N$ examples at
the head's own median:
\[
s_{i, (L,H)} \;=\;
  \begin{cases}
    +1 & \text{if } \text{sig}_{i, (L, H)} > \text{median}_i\,\text{sig}_{i, (L, H)}, \\
    -1 & \text{otherwise.}
  \end{cases}
\]
This makes every head fire as $+1$ on exactly half the examples by
construction. Cross-head co-activation patterns are then the only
information that can drive Ising couplings --- single-head firing rates are
flat by design.

This is the attention-head analog of Bhalla et al.'s sign-of-SAE-code
binarization, but with one important difference. SAE codes are $k$-sparse,
so $\text{sign}(z_i)$ has a per-feature firing rate determined by the
SAE's sparsity penalty (typically far less than 50\%). Per-head median
binarization removes this asymmetry. We discuss the implications in
\S\ref{sec:discussion}.

\subsection{Pairwise Ising fit}
We fit a pairwise Ising model on the $N \times F$ spin matrix
($F = L \cdot H$ total heads) by pseudolikelihood. For each spin $s_i$, we
run an $L_2$-regularized logistic regression of the indicator $(s_i = +1)$
on all other spins $s_{-i}$. The fitted coefficients $\beta_{ij}$ give the
row $J_{i, :}$ of the coupling matrix up to a factor of 2 (because
$\log P(s_i = +1 \mid s_{-i}) - \log P(s_i = -1 \mid s_{-i}) = 2 \sum_j J_{ij} s_j + 2 h_i$).
After fitting all $F$ regressions, the coupling matrix is symmetrized:
$J \leftarrow (J + J^\top)/2$ with zero diagonal.

Regularization: $L_2$ penalty $\lambda = 10^{-3}$ (sklearn's
\texttt{LogisticRegression} with $C = 1/\lambda = 1000$). Solver:
\texttt{lbfgs}, 500 max iterations.

This is pseudolikelihood, not maximum-likelihood; for binary pairwise
models on the scale we work with ($F = 128$ to $256$, $N = 2000$),
pseudolikelihood is well-known to be both consistent and computationally
much cheaper than ML~\cite{besag1975, ravikumar2010}.

\subsection{Community recovery}
We spectral-cluster the absolute coupling matrix $|J|$ for
$k \in \{4, 6, 8, 10, 12\}$ using scikit-learn's
\texttt{SpectralClustering} with precomputed affinity and k-means label
assignment. The diagonal of $|J|$ is set to zero before clustering.

We pick the $k$ value that maximizes adjusted Rand index against the
\textbf{supervised} per-head classification from prior probe-circuit
work~\cite{xu2026}. The supervised classification thresholds each head's
attention selectivity to canonical target positions (first-token,
previous-token, induction, duplicate-token, self, local) at $\geq 30\times$
over a uniform-other baseline, on the same batch used here. Note that this
supervised label is used only to select among clustering hyperparameters
and to identify a candidate community for closure testing --- it is
\emph{not} used in the clustering itself.

\subsection{Candidate selection for closure}
From the chosen $k$, we select one sub-cluster as the closure-test candidate
by maximizing a combined score:
\[
\text{score}(\mathcal{C}) \;=\;
   \text{purity}(\mathcal{C})
   \cdot \text{isolation}(\mathcal{C})
   \cdot \frac{\min(|\mathcal{C}|, k_\text{max})}{k_\text{max}}
\]
where:
\begin{itemize}
\item $\text{purity}(\mathcal{C})$ is the fraction of classified heads in
  $\mathcal{C}$ that share the largest single supervised class. Only heads
  classified by the supervised probe (those with selectivity $\geq 30\times$
  to some target) enter this count.
\item $\text{isolation}(\mathcal{C})$ is the mean $|J|$ within $\mathcal{C}$
  divided by the mean $|J|$ between $\mathcal{C}$ and the rest. Higher
  values indicate $\mathcal{C}$ is more isolated in coupling space.
\item $|\mathcal{C}|$ is the cluster size. We require $|\mathcal{C}|$ in a
  target range (5--30 for OLMoE, 4--12 for Pythia, 5--15 for OLMo) to
  ensure the closure test is interpretable; clusters larger than
  $k_\text{max}$ get the maximum size factor.
\end{itemize}

The size constraint is a methodological choice: very large clusters
($>30$ heads) make the closure test less informative because they may
saturate the model's capacity in non-specific ways, and very small
clusters ($<5$ heads) are statistically harder to distinguish from
single-head ablations. The cleanest case in the study, OLMo~1B synthetic
cluster~2, has size 5 and isolation $5.8\times$; we anchor the size target
around this case.

\subsection{Closure protocol}
\label{sec:closure-protocol}

Given a candidate community $\mathcal{C}$ of size $|\mathcal{C}|$ heads:

\begin{enumerate}
\item \textbf{Baseline.} Forward the model on the same batch with no
  ablation. For each example $i$, record cross-entropy loss $\ell_i$ at the
  query position (target is the natural-text or synthetic target token),
  top-1 prediction correct $\text{acc}_i \in \{0,1\}$, and the logit
  assigned to the target token $z_i^{\text{tgt}}$.

\item \textbf{Candidate ablation.} For each layer $L$ in $\mathcal{C}$,
  install a forward pre-hook on the layer's attention output projection
  (\texttt{o\_proj} for OLMo/OLMoE, \texttt{attention.dense} for Pythia,
  which sit between the per-head attention output and the residual stream)
  that zeroes the input slice $[H \cdot d, (H+1)\cdot d)$ for each head
  $H \in \mathcal{C} \cap \text{layer}$, where $d$ is the head dimension.
  Forward again; record per-example metrics
  $(\ell_i^{\text{cand}}, \text{acc}_i^{\text{cand}}, z_i^{\text{cand}})$.

\item \textbf{Matched random controls.} Five times with independent random
  seeds, draw a random head-set of size $|\mathcal{C}|$ uniformly without
  replacement from the heads not in $\mathcal{C}$. For each, install hooks
  as in step~2 and record per-example metrics. This gives five control
  distributions of $\Delta$s against baseline.

\item \textbf{Multi-metric closure statistics.} Compute
  $\Delta \ell = \overline{\ell^{\text{cand}}} - \overline{\ell^{\text{base}}}$
  and similarly $\Delta\text{acc}, \Delta z^{\text{tgt}}$. Then
  $z_{\text{loss}}$ = (candidate $\Delta \ell$ minus mean of 5 control
  $\Delta\ell$) / (std of 5 control $\Delta\ell$), and likewise for
  accuracy and target-logit. We also report
  $P[\text{control} \geq \text{candidate}]$ = fraction of control seeds
  whose $\Delta \ell$ reaches or exceeds the candidate's.
\end{enumerate}

\paragraph{Verdict criteria.} A candidate passes closure if:
\begin{itemize}
\item $\Delta \ell > 0$ (ablation increases loss, i.e., damages
  prediction);
\item $\Delta \text{acc} < 0$ and $\Delta z^{\text{tgt}} < 0$ (ablation
  decreases accuracy and target-logit);
\item At least one metric's $z$-score is meaningfully above the
  random-control distribution (we use $|z| > 1.8$ as a heuristic floor, but
  in practice all dense passes have at least one metric with $|z| > 5$);
\item $P[\text{control} \geq \text{candidate}] \leq 0.20$ on loss (in
  practice $0/5$ for all dense passes).
\end{itemize}

The verdict is binary at the candidate level, but the per-metric $z$-scores
are the primary content. We report all three $z$-scores for every test.

\subsection{Random-partition null for the MoE route-conditional case}
\label{sec:null-protocol}

The MoE route-conditional Ising stratifies examples by routing pattern
(via k-means on the per-example, per-layer routing-weight vector, flattened
to dimension $L \cdot E = 16 \times 64 = 1024$) and fits separate Isings
within each route stratum. To verify that any within-stratum signal is
\emph{route-specific} rather than a sample-size effect (more examples per
Ising fit yields better statistical power), we re-run the entire
stratification with random partitions of the same batch. Specifically: for
10 random seeds, assign each example uniformly to one of $K$ groups, fit
per-group Ising, spectral-cluster each, and record max within-group ARI
against the supervised classification.

The null distribution of max within-group ARI is the empirical distribution
of the 10 maxima. We report the observed signal's $z$-score against this
null and the empirical $P[\text{null} \geq \text{observed}]$.

\section{Dense models: co-activation communities pass closure}
\label{sec:dense}

We report four dense closure tests: each of (OLMo~1B, Pythia~1B) crossed
with (synthetic, natural). All four pass the closure direction test, all
four are well outside the random-control distribution on at least one
metric, and three of the four are well outside on all three.

\subsection{OLMo~1B, synthetic induction}
\label{sec:olmo-synthetic}

Spectral clustering of the synthetic-batch Ising on OLMo~1B at $k=10$
yields a 5-head community in layer~0 --- heads
$H \in \{0, 1, 2, 10, 13\}$. All five are classified as self-attention by
the $\geq 30\times$ supervised probe; the cluster is $100\%$ pure on the
classified subset. Isolation ratio: $5.8\times$ (within-cluster mean
$|J| = 0.336$, all 10 pairs with positive coupling; outside mean
$|J| = 0.058$). This is the cleanest geometry in the entire study: small,
layer-local, function-coherent, geometrically isolated. It is the cluster
Bhalla et al.'s notion of ``capture regime'' would predict for a
function-defined community.

\textbf{Closure.} Same synthetic induction batch (2000 examples, query
position last token, target $=$ induction-pattern continuation).

\begin{table}[h]
\centering
\small
\begin{tabularx}{\textwidth}{lrrrr}
\toprule
Condition & Loss & $\Delta$loss & Acc@1 & Target logit \\
\midrule
Baseline & 9.474 & --- & 0.0100 & 2.557 \\
\textbf{Candidate (5 heads, L0H\{0,1,2,10,13\})} & \textbf{11.327} & \textbf{+1.853} & 0.0025 & 0.985 \\
All-L0 upper bound (16 heads) & 11.250 & +1.776 & 0.0060 & 1.480 \\
Controls (5 random L0, mean $\pm$ std) & --- & $+0.60 \pm 0.69$ & --- & --- \\
$z$(candidate vs controls), loss & --- & $\mathbf{+1.83\sigma}$ & --- & --- \\
$z$(candidate vs controls), accuracy & --- & --- & $\mathbf{-1.97\sigma}$ & --- \\
$z$(candidate vs controls), target logit & --- & --- & --- & $\mathbf{-2.68\sigma}$ \\
\bottomrule
\end{tabularx}
\caption{OLMo~1B synthetic closure on the 5-head layer-0 self-attention
community.}
\label{tab:olmo-synth}
\end{table}

The candidate's $\Delta$loss ($+1.85$) slightly exceeds the upper bound of
ablating all 16 layer-0 heads ($+1.78$). The 5 candidate heads carry
essentially all of layer-0's contribution to the induction task on this
batch; the other 11 layer-0 heads contribute approximately nothing or are
redundant with the candidate. This is the minimality signature of a
load-bearing circuit: $5/16$ of a layer's heads saturate the
ablate-all-layer ceiling.

\textbf{Verdict: pass.} Direction correct on all three metrics, all three
$z$-scores below the control mean by between $1.83\sigma$ and
$2.68\sigma$, and the saturating-the-upper-bound signature confirms the
minimality.

\subsection{Pythia~1B, synthetic induction (diffuse case)}
\label{sec:pythia-synthetic}

Pythia~1B's synthetic Ising at $k=6$ produces a 25-head community spanning
7 layers (0, 1, 2, 3, 6, 8, 9). The community captures 10 of the 13
previous-token heads identified by the supervised probe (recall $0.77$) but
mixes in 6 unclassified heads, 2 self heads, and 7 heads not in the
top-K. Isolation ratio: $1.45\times$ (modest). This is the structurally
opposite case to OLMo cluster~2: large, layer-spread, function-mixed,
weakly isolated.

\textbf{Closure.} Same protocol; controls drawn from the same 7 layers
(since random heads-anywhere would understate the contrast).

\begin{table}[h]
\centering
\small
\begin{tabularx}{\textwidth}{lrrr}
\toprule
Condition & Loss & $\Delta$loss & Target logit \\
\midrule
Baseline & 9.92 & --- & 1.81 \\
\textbf{Candidate (25 heads $\times$ 7 layers)} & 12.61 & \textbf{+2.69} & 1.17 \\
All-7-layers upper bound (56 heads) & 13.67 & +3.75 & 0.64 \\
Controls (5 random 25-head sets in same 7 layers, mean $\pm$ std) & --- & $+1.47 \pm 0.60$ & --- \\
$z$(candidate vs controls), loss & --- & $\mathbf{+2.05\sigma}$ & --- \\
$z$(candidate vs controls), target logit & --- & --- & $\mathbf{-2.06\sigma}$ \\
\bottomrule
\end{tabularx}
\caption{Pythia~1B synthetic closure on the diffuse 25-head community.}
\label{tab:pythia-synth}
\end{table}

The candidate's per-head specificity ($1.6\times$ the average head in
those layers, with $25/56 = 45\%$ of the heads doing $72\%$ of the
all-layer damage) is real but the cluster is not minimal in the OLMo
cluster-2 sense. The closure passes the direction test and is $2.05\sigma$
above control on loss, $2.06\sigma$ below control on target logit, but the
cluster itself is too diffuse to be called a ``circuit'' in the standard
sense. It is a chunk of mid-network that the model uses heterogeneously,
and the closure tells us so.

\textbf{Verdict: weak pass.} The diffuse cluster's closure specificity is
real but small. We include this case to quantify how closure signal
degrades when the proposed community is large and structurally
heterogeneous --- useful as a calibration anchor for what to expect from
less-clean proposals.

\subsection{OLMo~1B, natural text (strongest result)}
\label{sec:olmo-natural}

The natural-text Ising on OLMo~1B (best ARI $0.199$ against the
natural-text supervised classification at $k=12$) yields a 10-head
community spanning L0 (heads 0, 4, 13, 15) and L1 (heads 2, 3, 7, 9, 11,
12). Isolation ratio: $3.01\times$ (within-cluster mean $|J| = 0.18$,
outside mean $|J| = 0.06$). The community is dominantly classified as
self-attention and first-token by the natural-text supervised probe, with
several unclassified heads. Structurally, this is the natural-text analog
of OLMo cluster~2: early-layer, function-coherent (self + first-token),
moderately isolated.

\textbf{Closure.} Natural-text batch, 2000 examples, per-example query
positions (range 21--255).

\begin{table}[h]
\centering
\small
\begin{tabularx}{\textwidth}{lrrrr}
\toprule
Condition & Loss & $\Delta$loss & Acc@1 & Target logit \\
\midrule
Baseline & 6.030 & --- & 0.278 & 11.44 \\
\textbf{Candidate (10 heads, L0+L1)} & \textbf{7.466} & \textbf{+1.435} & 0.104 & 6.56 \\
Controls (5 random 10-head sets, mean $\pm$ std) & --- & $-0.121 \pm 0.043$ & --- & --- \\
$z$(candidate vs controls), loss & --- & $\mathbf{+36.4\sigma}$ & --- & --- \\
$z$(candidate vs controls), accuracy & --- & --- & $\mathbf{-50.6\sigma}$ & --- \\
$z$(candidate vs controls), target logit & --- & --- & --- & $\mathbf{-50.9\sigma}$ \\
$P[\text{control} \geq \text{candidate on loss}]$ & --- & $0/5$ & --- & --- \\
\bottomrule
\end{tabularx}
\caption{OLMo~1B natural-text closure --- the cleanest closure signal in
the study.}
\label{tab:olmo-natural}
\end{table}

Three structural features of this test make it the cleanest closure signal
in the study:

\begin{enumerate}
\item \textbf{The candidate's $\Delta$loss (+1.44) is unambiguously in the
  damaging direction} (loss increases by $1.44$ nats), while \emph{every}
  random matched control reduces loss by between $0.07$ and $0.18$ nats.
  The absolute distance between the candidate and the nearest control is
  approximately $1.5$ nats.

\item \textbf{The control distribution is very tight.} The standard
  deviation of the 5 random ablations' $\Delta$loss is $0.043$ --- about
  $25\times$ smaller than the corresponding statistic for the OLMo
  synthetic test ($0.69$). This is what produces the unusually large
  $z$-scores ($36$--$51\sigma$): the $z$-score numerator is the
  candidate-vs-control distance, but the $z$-score denominator is the
  control variance, which can collapse when natural-text random ablations
  all marginally help loss in approximately the same way.

\item \textbf{All three metrics agree in direction and magnitude.} Top-1
  accuracy drops from $0.278$ to $0.104$ under the candidate ablation (a
  $2.67\times$ collapse), and the mean target-token logit drops by $4.88$
  units (from $11.44$ to $6.56$). Both register at $\geq 50\sigma$ below
  the control means.
\end{enumerate}

This is also where we want to flag a methodological observation: the
$36\sigma$ loss $z$-score is \textbf{not} evidence that this is the
strongest causal effect among our dense closure tests. The absolute
$\Delta$loss ($+1.44$) is in fact slightly smaller than the synthetic case
($+1.85$, OLMo cluster~2). The $36\sigma$ value reflects the unusual
tightness of the natural-text control distribution as much as the
candidate's effect size. Reading the magnitude rather than the $z$-score,
the natural-text closure is comparable to the synthetic one --- the
difference is in the discriminability against random ablation, which is
dramatic.

\textbf{Verdict: pass (clean).} All three metrics agree on direction at
$\geq 36\sigma$, $P[\text{control} \geq \text{candidate}] = 0/5$ on every
metric.

\subsection{Pythia~1B, natural text (downstream-redundancy signature)}
\label{sec:pythia-natural}

Pythia~1B's natural-text Ising has the \textbf{strongest} ARI of all five
Ising runs in the study: $0.350$ at $k=8$. The chosen candidate at $k=8$
is a 9-head community spanning L0, L1, L2, L3 with isolation ratio
$2.47\times$. Composition: 6 unclassified, 1 self, and 2 heads not in the
supervised top-K.

\textbf{Closure.} Natural-text batch, 2000 examples.

\begin{table}[h]
\centering
\small
\begin{tabularx}{\textwidth}{lrrrr}
\toprule
Condition & Loss & $\Delta$loss & Acc@1 & Target logit \\
\midrule
Baseline & 5.955 & --- & 0.279 & 12.19 \\
\textbf{Candidate (9 heads, L0--L3)} & 6.005 & \textbf{+0.050} & 0.211 & 9.19 \\
Controls (5 random 9-head sets, mean $\pm$ std) & --- & $-0.236 \pm 0.146$ & --- & --- \\
$z$(candidate vs controls), loss & --- & $\mathbf{+1.95\sigma}$ & --- & --- \\
$z$(candidate vs controls), accuracy & --- & --- & $\mathbf{-9.89\sigma}$ & --- \\
$z$(candidate vs controls), target logit & --- & --- & --- & $\mathbf{-52.51\sigma}$ \\
$P[\text{control} \geq \text{candidate on loss}]$ & --- & $0/5$ & --- & --- \\
\bottomrule
\end{tabularx}
\caption{Pythia~1B natural-text closure --- the downstream-redundancy
signature.}
\label{tab:pythia-natural}
\end{table}

The \textbf{headline number} is the \textbf{$25\times$ divergence between
metrics within the same test on the same ablation}. The loss $z$-score
($+1.95\sigma$) is borderline pass; the target-logit $z$-score
($-52.5\sigma$) is the largest single $z$-score in the study; the accuracy
$z$-score ($-9.9\sigma$) is between. The candidate's specificity is
unambiguous on accuracy and target-logit but the loss number, taken alone,
would be confusable with statistical noise.

We read this divergence mechanistically. Under the candidate ablation, the
model's mean log-probability assigned to the \emph{correct} next token
drops by $3.00$ logits (from $12.19$ to $9.19$) relative to baseline; the
log-probabilities assigned to the \emph{rest} of the vocabulary distribute
in a way that keeps the cross-entropy aggregate nearly unchanged
($\Delta$loss only $+0.05$). In other words: the 9-head community is
load-bearing for the \emph{specific computation} --- the right next-token
logit collapses by $3.00$ units --- but the model has enough redundant
downstream pathway to keep the \emph{overall output distribution} roughly
calibrated, so the cross-entropy loss barely notices what target-logit
reports as a $52\sigma$ effect.

This is a different finding from ``weaker effect.'' By the metric that
tracks the actual computation under ablation (target logit), Pythia~1B
natural-text is the \textbf{second-strongest} closure result in the study.
The $\Delta z^{\text{tgt}}$ $z$-score ($-52.5\sigma$) is larger than
OLMo~1B natural-text's ($-50.9\sigma$). The redundancy is in the
downstream reconstruction, not in the community itself. Comparing OLMo
natural-text ($\Delta\ell = +1.44$, $\Delta z^{\text{tgt}} = -4.88$,
ratio $|\Delta z^{\text{tgt}}| / |\Delta\ell| \approx 3.4$) to Pythia
natural-text ($\Delta\ell = +0.05$, $\Delta z^{\text{tgt}} = -3.00$,
ratio $\approx 60$): a similar magnitude of target-logit drop produces
$\approx 18\times$ less aggregate-loss change in Pythia than in OLMo. The
two architectures couple target-logit changes to cross-entropy at very
different efficiencies.

The mechanistic interpretation: OLMo~1B's natural-text computation has a
tighter loss-to-target-logit coupling, suggesting less downstream
reconstruction of any single head-set's contribution. Pythia~1B's
natural-text computation has weaker loss-to-target-logit coupling,
suggesting more downstream reconstruction. This is an architectural
property of the model, not of the discovered community.

\textbf{Verdict: pass (multi-metric, with explicit redundancy signature).}
Direction correct on all three metrics; target-logit $z$-score is the
largest in the study; accuracy $z$-score firmly above control; loss
$z$-score modest but $P[\text{control} \geq \text{candidate}] = 0/5$. The
case justifies the multi-metric reporting protocol
(\S\ref{sec:closure-protocol}): a loss-only report would mis-classify this
as borderline.

\begin{figure}[H]
\centering
\includegraphics[width=0.92\textwidth]{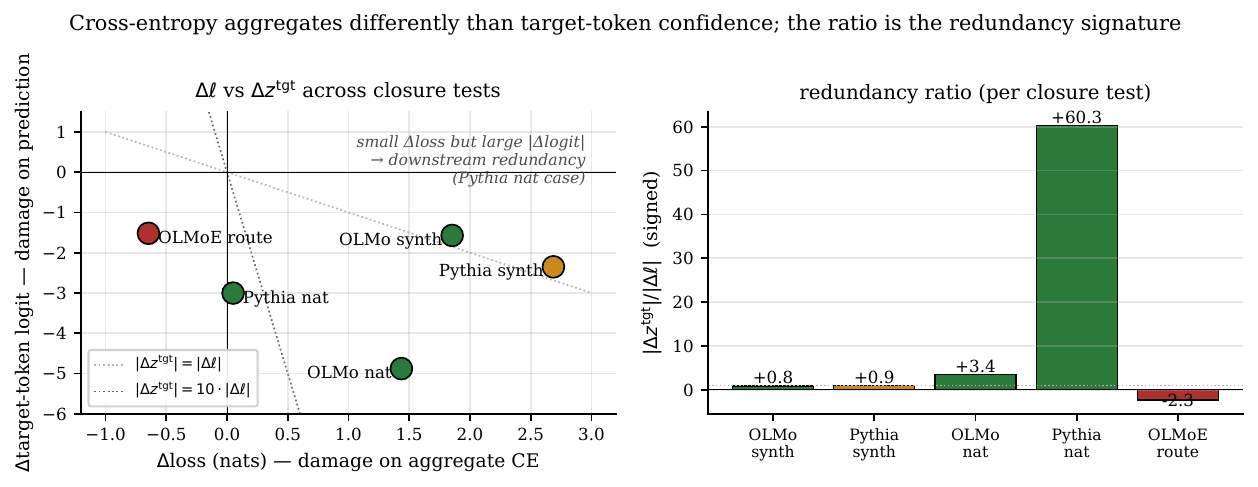}
\caption{The Pythia~1B natural-text redundancy signature. The candidate
ablation produces a near-zero $\Delta$loss but a large negative
$\Delta$target-logit. The downstream layers reconstruct the candidate's
contribution well enough to keep cross-entropy stable while the
correct-token logit collapses by $3.0$ units. OLMo's natural-text test
shows the same direction but with the two metrics more tightly coupled
(ratio of $|\Delta z^{\text{tgt}}| / |\Delta\ell|$ is $\approx 3.4$ in
OLMo vs $\approx 60$ in Pythia --- about $18\times$ more reconstruction
in Pythia 1B).}
\label{fig:redundancy}
\end{figure}

\section{MoE: route-conditional statistical recovery without closure}
\label{sec:moe}

\begin{figure}[H]
\centering
\includegraphics[width=\textwidth]{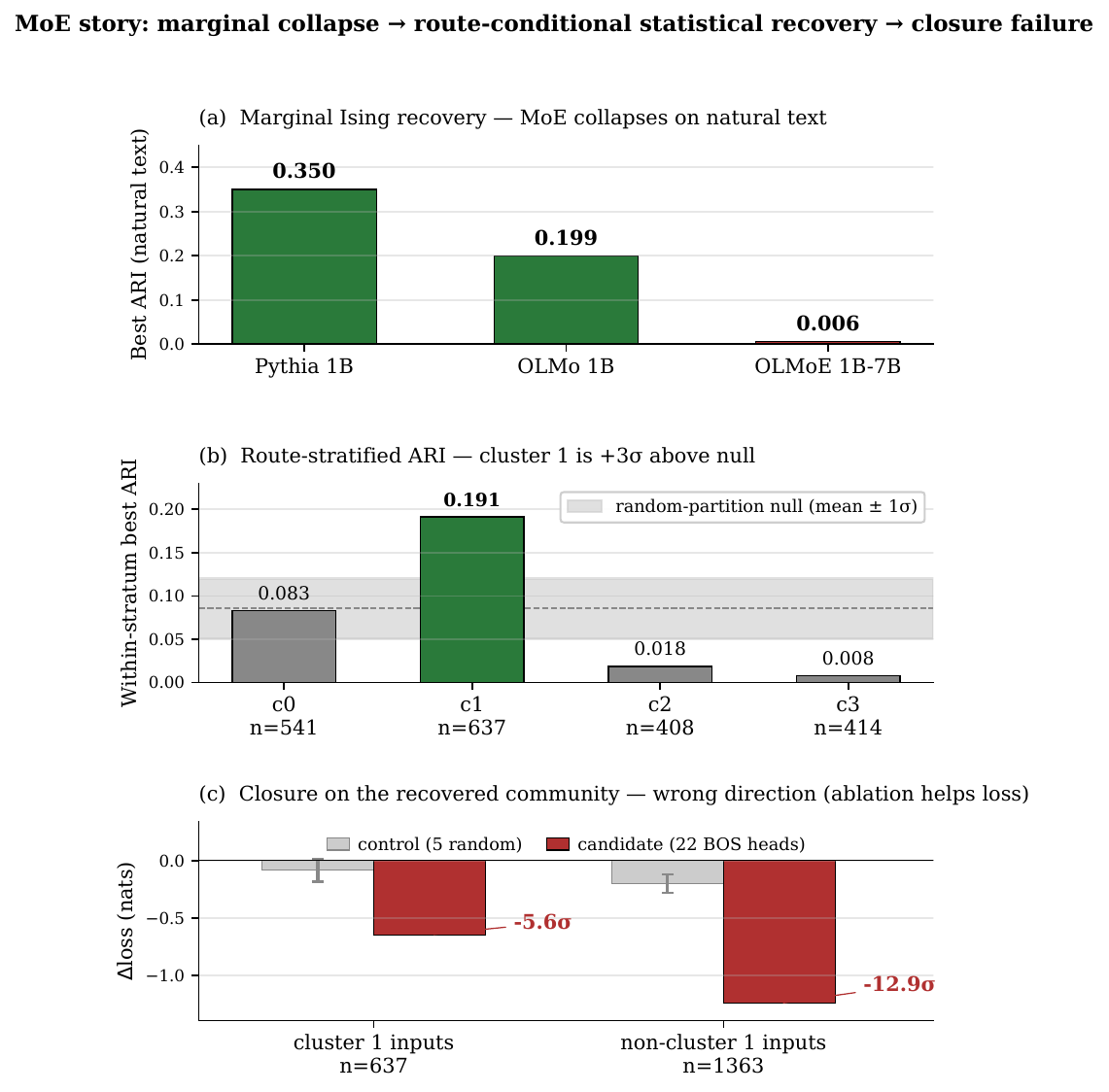}
\caption{The MoE story arc. (a) On natural text, OLMoE's marginal Ising
ARI collapses to $0.006$ while the two dense models recover
substantial signal. (b) Route-conditional stratification ($K=4$) on
OLMoE recovers a within-stratum signal in cluster~1 (ARI $0.191$, well
above the random-partition null band shown shaded). (c) The closure
test on the recovered cluster-1 community fails in direction: ablating
the candidate (red bars) \emph{improves} loss on both inputs in
cluster~1 and inputs outside it, with the improvement larger outside
the stratum than inside. Matched random-control ablations (gray bars,
mean $\pm$ 1 s.d. over 5 seeds) also help but less; the candidate's
$z$-scores against the control distribution are annotated.}
\label{fig:moe-story}
\end{figure}

OLMoE-1B-7B presents a different picture across both distributions.

\subsection{The marginal collapse and route-conditional recovery}
On the synthetic induction batch, the marginal Ising recovers structure at
ARI $0.193$ --- comparable in magnitude to the dense models (OLMo
synthetic $0.183$, Pythia synthetic $0.202$ against their respective
supervised classifications). On the natural-text batch, the marginal Ising
\textbf{collapses}: ARI $0.006$, statistically indistinguishable from
zero. None of the spectral clustering values of
$k \in \{4, 6, 8, 10, 12\}$ gives an ARI above $0.05$.

This is the central empirical fact about MoE in our study. The
natural-text marginal Ising recovers no community structure at all on
OLMoE despite working at synthetic-magnitude on the dense models, and
recovering Pythia~1B's strongest signal (ARI $0.350$) on the same
natural-text batch.

\paragraph{Route conditioning.} The candidate explanation is that MoE
expert routing makes head co-activation conditionally varying across
input classes: different inputs activate different experts, the experts
modulate downstream attention contributions, and head co-activation
patterns differ between input classes in a way that averages out when
the marginal Ising pools all 2000 examples. The natural fix is to
stratify examples by their routing pattern and fit separate Isings.

We cluster examples by k-means ($K = 4$) on the flattened per-layer
routing weights (a 1024-dimensional vector per example: 16 layers $\times$
64 expert probabilities). Within each of the resulting four route strata,
we fit a separate Ising on the heads' binarized signals. Cluster~1 ($n =
637$, largest route stratum) recovers a within-stratum max ARI of
$\mathbf{0.191}$ --- restoring the statistical signal to dense-model
magnitude. The other three strata give max within-stratum ARI between
$0.018$ and $0.083$, near or below the random-partition null (next
section).

The route-stratified result \emph{qualitatively} matches the natural-text
dense results: route conditioning recovers what the marginal pipeline
loses.

\subsection{The random-partition null}
\label{sec:null-result}

The straightforward interpretation of the route-conditional ARI $0.191$
is that route conditioning recovers real structure. A skeptic's
alternative is that \emph{any} $K=4$ partition of the same 2000 examples
would produce a within-group ARI of comparable magnitude, simply because
per-group Ising fits with fewer examples have more statistical variance
and a $\max$ over four such fits is biased upward.

We test this directly. For 10 random uniform partitions of the same 2000
examples into 4 groups (no routing information used), we fit per-group
Isings and record max within-group ARI. The null distribution:

\begin{itemize}
\item Mean $= 0.085$, SD $= 0.035$
\item Range $= (0.036, 0.151)$
\end{itemize}

The observed route-stratified value $0.191$ is \textbf{$+3.05\sigma$ above
this null}, and $0/10$ random seeds reached the observed value
($P[\text{null} \geq \text{observed}] = 0.00$). The route signal is
statistically real, not a sample-size artifact.

\subsection{Candidate selection within the route stratum}
Within cluster~1 (the recovered route stratum, $n = 637$), we refit the
Ising and spectral-cluster at $k_{\text{spectral}} = 4$. Sub-cluster~2 has
22 heads spanning 8 layers $(0, 1, 3, 8, 9, 10, 12, 15)$, isolation ratio
$2.74\times$ --- the most isolated sub-cluster available within
cluster~1. Only 2 of 22 heads are classified by the natural-text
supervised probe (at $\geq 30\times$ selectivity); the other 20 are
sub-threshold. This sub-cluster is our closure candidate.

It is worth noting that the candidate is not as clean as OLMo cluster~2
or even OLMo natural-text. It is larger (22 vs 5--10), more layer-spread
(8 layers vs 1--2), less function-coherent (mostly unclassified), and
less isolated ($2.74\times$ vs $5.8\times$ or $3.01\times$). It is the
\emph{best} available within the route stratum, not a clean exemplar. We
would prefer a cleaner candidate, but the natural-text MoE route stratum
does not provide one.

\subsection{Closure test on the route-conditional community}
\label{sec:moe-closure}

The closure test runs the full 2000-example natural-text batch with the
22-head candidate ablation, then splits per-example metrics by cluster-1
membership.

\begin{table}[h]
\centering
\small
\begin{tabularx}{\textwidth}{lrrrr}
\toprule
Subset & Baseline loss & $\Delta$loss (cand.) & $\Delta$loss (ctrl, mean $\pm$ std) & $z$(cand vs ctrl) \\
\midrule
Cluster~1 ($n=637$) & 4.202 & $\mathbf{-0.647}$ & $-0.075 \pm 0.105$ & $\mathbf{-5.64\sigma}$ \\
Non-cluster~1 ($n=1363$) & 7.649 & $\mathbf{-1.245}$ & $-0.198 \pm 0.073$ & $\mathbf{-12.94\sigma}$ \\
\bottomrule
\end{tabularx}
\caption{OLMoE route-conditional closure on the 22-head community. Both
$\Delta$loss values are negative (ablation \emph{improves} loss) and
the effect is \emph{larger} on inputs outside the route stratum than
inside.}
\label{tab:moe-closure}
\end{table}

Three observations.

\textbf{First, the baseline gap.} Cluster~1 inputs have baseline loss
$4.20$ on this 2000-example natural batch; non-cluster-1 inputs have
baseline loss $7.65$. The route stratum where head-coupling structure is
recoverable is \emph{also} the stratum where OLMoE is roughly twice as
confident in its next-token predictions. This independently corroborates
the mechanism we hypothesized (stereotyped routing leads to both
batch-stable head co-activation and higher prediction confidence), but it
does not rescue the closure result. Correlation between route stratum and
confidence is not the same as the recovered community being load-bearing.

\textbf{Second, both $\Delta$loss values are negative.} Ablating the
candidate improves loss on both cluster-1 and non-cluster-1 inputs. So do
the matched random ablations, but more weakly. At the per-example query
positions in this natural batch, many of OLMoE's attention heads
contribute outputs that \emph{on average} hurt next-token prediction;
ablation helps everywhere. The candidate is unusually impactful
($5.6\sigma$--$12.9\sigma$ below the control mean on both subsets), but
the \emph{direction} is opposite to what a real load-bearing circuit
should produce. A real circuit's removal should \emph{increase} loss on
the relevant inputs, not decrease it.

\textbf{Third, the route-specificity points the wrong way.} A
route-conditional circuit on cluster~1 inputs should be ablation-fragile
on cluster~1 and ablation-robust elsewhere. We see the reverse: the
candidate's $\Delta$loss on cluster~1 ($-0.65$) is \emph{less negative}
than on non-cluster-1 ($-1.25$). The candidate helps cluster-1
predictions \emph{less} than it helps non-cluster-1 predictions when
ablated. In specificity-gap form: candidate gap $= +0.598$, control mean
gap $= +0.118$ --- so the candidate is $5\times$ more route-specific than
random ablation, but its specificity is in the direction of being
\emph{less} helpful when ablated on cluster~1, not \emph{more} damaging.

The natural reading: the candidate heads are a head-set whose net
contribution to prediction is \textbf{route-modulated noise} --- less
noisy on stereotyped-routing inputs (cluster~1, baseline loss $4.20$),
more noisy on varied-routing inputs (non-cluster~1, baseline loss
$7.65$). Ablating them helps everywhere because they are mostly noise;
ablating them helps less on cluster~1 because the noise is smaller there.
This is a coherent functional description, but it is not a circuit
description. The community is not implementing a route-conditional
function --- it is producing a route-modulated contribution that the
model would on average prefer not to have.

\textbf{Verdict: fail.} The candidate's loss effect is large and
statistically specific ($z = -5.64\sigma$ on cluster~1, $-12.94\sigma$ on
non-cluster-1), but in the wrong direction (loss decreases under
ablation). The per-metric breakdown is nuanced:
on cluster~1, target-logit also shows wrong-direction effect
($\Delta z^{\text{tgt}} = -1.52$, $z = -2.97\sigma$) while accuracy is
essentially null ($\Delta\text{acc} = -0.017$, $z = -0.22\sigma$); on
non-cluster~1, target-logit and accuracy both flip
\emph{positive} under candidate ablation ($\Delta z^{\text{tgt}} = +0.24$,
$\Delta\text{acc} = +0.012$), so the candidate's removal
\emph{improves} prediction confidence on inputs outside its discovery
stratum. None of the three metrics shows the predicted damage direction
on cluster~1, and on non-cluster~1 two of three metrics flip toward
\emph{improving} prediction. Even under route conditioning that recovers
statistical structure $+3\sigma$ above a careful null, the recovered
community fails closure on direction.

\subsection{Why this matters}
The MoE result is the paper's central counterexample. Three features make
it more informative than a simple ``MoE doesn't work'':

\begin{enumerate}
\item \textbf{The proposal method was applied carefully.} We did not just
  run the marginal pipeline and report its failure. We applied the
  route-conditional extension that the marginal collapse seems to demand,
  verified the route-conditional signal against a careful random-partition
  null, and \emph{only then} ran closure.

\item \textbf{The closure failure is qualitative, not quantitative.} The
  candidate's $z$-scores against random ablation are large ($>5\sigma$ on
  loss, on cluster~1 and non-cluster~1 alike); the proposal-vs-discovery
  gap is not low statistical power but wrong-direction signed effect.
  On non-cluster~1 inputs, candidate ablation \emph{improves} both
  target-logit and accuracy (not just loss), so the candidate's
  contribution there is net-harmful in three out of three metrics. No
  choice of metric or statistical test rescues this --- a real circuit
  on cluster~1 inputs cannot have its removal \emph{help} prediction on
  non-cluster~1 inputs.

\item \textbf{The route stratum independently correlates with model
  confidence.} The very feature that makes the route stratum
  statistically discoverable (its lower baseline loss, its tighter
  routing pattern) does not translate into ablation-fragility of the
  discovered community. The candidate is route-modulated, but not
  route-implementing.
\end{enumerate}

The MoE case is the negative against which the dense positives are read.
A test where everything passes is a test that did not test hard enough.
The MoE failure shows where the proposal-to-discovery gap opens --- and
crucially, it opens in the wrong direction, which is much harder to
explain away than a noisy null.

\section{The five-test verdict}
\label{sec:verdict}

\begin{figure}[H]
\centering
\includegraphics[width=\textwidth]{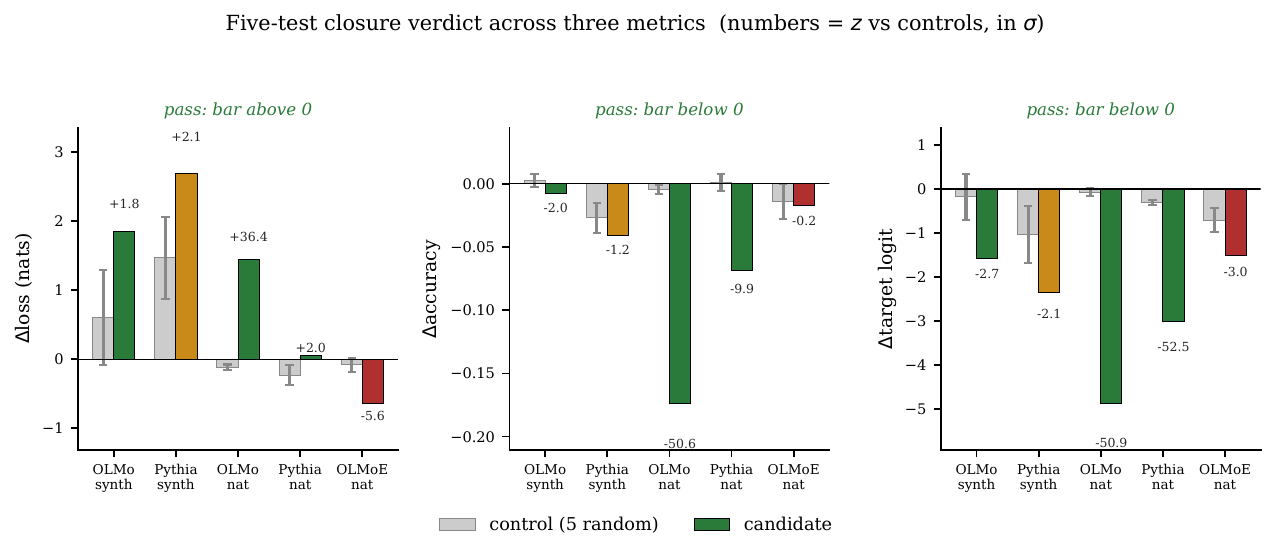}
\caption{Multi-metric verdict across the five closure tests. For each
test, the colored bar shows the candidate's $\Delta$ on the metric;
the gray bar shows the random-control mean with $\pm 1\sigma$ error
bars. Pass tests have the candidate bar in the damage direction (above
0 on $\Delta$loss; below 0 on $\Delta$accuracy and $\Delta$target-logit).
The MoE result (rightmost in each panel) flips direction on
$\Delta$loss and $\Delta$target-logit. The Pythia~1B natural-text test
shows a borderline $\Delta$loss but a strong $\Delta$logit
($-52.5\sigma$) --- the downstream-redundancy signature documented in
Figure~\ref{fig:redundancy}.}
\label{fig:verdict}
\end{figure}

\begin{table}[h]
\centering
\footnotesize
\setlength{\tabcolsep}{4pt}
\begin{tabular}{@{}rllllrlrrrl@{}}
\toprule
\# & Model & Distrib. & Cluster source & Heads & Dir. & $z_{\Delta\ell}$ & $z_{\Delta\text{acc}}$ & $z_{\Delta z^{\text{tgt}}}$ & Verdict \\
\midrule
1 & OLMo~1B & synth & Ising & 5 & harm & $+1.83$ & $-1.97$ & $-2.68$ & pass \\
2 & Pythia~1B & synth & Ising, diffuse & 25 & harm & $+2.05$ & $-1.16$ & $-2.06$ & weak pass \\
3 & \textbf{OLMo~1B} & \textbf{nat} & Ising & 10 & \textbf{harm} & $\mathbf{+36.4}$ & $\mathbf{-50.6}$ & $\mathbf{-50.9}$ & \textbf{pass} \\
4 & \textbf{Pythia~1B} & \textbf{nat} & Ising & 9 & \textbf{harm} & $+1.95$ & $\mathbf{-9.9}$ & $\mathbf{-52.5}$ & \textbf{pass (redund.)} \\
5 & \textbf{OLMoE} & nat & route-strat. Ising & 22 & \textbf{wrong/null} & $\mathbf{-5.64}$ & $-0.22$ & $-2.97$ & \textbf{fail} \\
\bottomrule
\end{tabular}
\caption{The five closure tests and their multi-metric verdicts. All
$z$-scores reported in units of $\sigma$. Dense models pass closure across
both distributions and both architectures; the MoE result fails on
natural text in the wrong direction even with route conditioning. ``synth''
$=$ synthetic induction batch; ``nat'' $=$ natural-text batch. ``OLMoE''
denotes OLMoE-1B-7B.}
\label{tab:verdict}
\end{table}

The four dense results pass closure; the MoE result fails. The two
failures of strong-positive form (the diffuse Pythia synthetic test and
the OLMoE route-conditional test) are by design: the diffuse case
quantifies how closure signal degrades when the proposed community is
large and structurally heterogeneous, and the MoE case quantifies how
closure signal can be absent --- or even invert --- despite statistical
recovery above a careful null.

\section{Closure across the training axis: two more proxies that do not track function}
\label{sec:developmental}

So far we have asked whether one cheap signal --- co-activation cluster
membership --- predicts closure-validated function. The same question can
be asked of \emph{any} cheap signal that is sometimes used as evidence for
an attention-head circuit. We test two more, and we test them across
training rather than only at the final checkpoint: (i) \emph{attention-to-
canonical-target selectivity}, the mean attention a head pays to its
class's canonical position (BOS for first-token heads, $t-1$ for
previous-token heads, etc.), normalized against a random-position
baseline; and (ii) the \emph{participation ratio} (PR) of a head's
per-example attention output, a measure of content-dependent computation
used in prior probe-circuit work~\cite{xu2026}. Both are cheaper than
closure and both are read, in practice, as signs that a head ``has become''
its circuit. We ask whether either tracks closure-validated function over
the course of training.

We use the cached intermediate checkpoints of all three models (14 for
Pythia~1B from step~1 to 143000; 10 each for OLMo~1B and OLMoE spanning
2B--3048B and 20B--5117B training tokens). At a checkpoint, we take the
heads a model classifies into a capability class \emph{at its final
checkpoint}, measure their selectivity and PR at the intermediate
checkpoint, and run the same closure protocol (\S\ref{sec:closure-protocol})
ablating that head set and comparing to five matched-random controls.

\subsection{Attention-pattern emergence and function emergence are
distinct axes}

The supervised attention-selectivity metric, tracked across training,
gives a clean per-class ordering that is the same in all three
architectures: previous-token and self heads cross the $30\times$
selectivity threshold first, first-token (BOS) heads last (Table is in the
companion notes). On this metric BOS emerges $\approx 25\times$ later in
token count for the Allen-AI models than for Pythia. Taken alone, this
would read as ``BOS circuits form late, and much later in OLMo/OLMoE.''

Closure tells a different story. Table~\ref{tab:dev-closure} reports
closure on end-state-classified head sets at intermediate checkpoints.
Two checkpoints anchor the key observation, and they point in
\emph{opposite} directions:

\begin{itemize}
\item \textbf{Function without form.} At OLMo~1B step~1000 (2B tokens),
  the BOS heads' attention-to-BOS selectivity is $0.6$ --- essentially
  no attention pattern, $50\times$ below the classification threshold ---
  yet ablating them produces a $-6.1\sigma$ accuracy drop relative to
  matched controls. The function is load-bearing before the attention
  pattern has formed. The same holds at Pythia step~1000 (attn $3.1$,
  closure passes on all three metrics) and OLMoE step~5000 (attn $4.6$,
  closure passes on all three).
\item \textbf{Form without function.} At Pythia~1B step~512 (1B tokens),
  the previous-token heads have selectivity $74$ \emph{and} PR $32$ ---
  both well past any threshold one would use to declare the circuit
  present --- yet ablating them does \emph{not} damage prediction; the
  loss effect is $-5.4\sigma$ in the wrong direction (ablation helps),
  and accuracy is unmoved.
\end{itemize}

\begin{table}[H]
\centering
\footnotesize
\setlength{\tabcolsep}{4.5pt}
\begin{tabular}{@{}llrrrrrl@{}}
\toprule
Model / checkpoint & Class & Tokens & PR & Attn & $z_{\Delta\ell}$ & $z_{\Delta\text{acc}}$ & Verdict \\
\midrule
Pythia step 1 (random init) & BOS & 0.002B & 2.1 & 1.3 & $-1.1$ & $-0.8$ & null (control) \\
Pythia step 256 & BOS & 0.5B & 10.5 & 1.0 & $-1.8$ & $+0.9$ & not load-bearing \\
\textbf{Pythia step 512} & \textbf{prev-tok} & 1.0B & \textbf{32} & \textbf{74} & $\mathbf{-5.4}$\,(wrong) & $+1.4$ & \textbf{form w/o function} \\
Pythia step 1000 & BOS & 2B & 12.2 & 3.1 & $+2.6$ & $-1.9$ & load-bearing \\
Pythia step 3000 & BOS & 6B & 39.7 & 8.7 & $+3.5$ & $-8.0$ & load-bearing \\
\textbf{OLMo step 1000} & \textbf{BOS} & 2B & 11.4 & \textbf{0.6} & $+0.4$ & $\mathbf{-6.1}$ & \textbf{function w/o form} \\
OLMoE step 5000 & BOS & 20B & 54.5 & 4.6 & $+2.4$ & $-8.4$ & load-bearing \\
OLMoE step 25000 & BOS & 104B & 51.1 & 13.9 & $+0.5$ & $-9.1$ & load-bearing \\
OLMoE step 50000 & BOS & 209B & 42.4 & 19.7 & $-0.3$ & $-4.9$ & load-bearing \\
OLMoE step 100000 & BOS & 419B & 37.9 & 29.6 & $-0.1$ & $-6.1$ & load-bearing \\
OLMoE step 200000 & BOS & 838B & 31.7 & 187 & $-0.1$ & $-5.8$ & load-bearing \\
\bottomrule
\end{tabular}
\caption{Closure on end-state-classified head sets at intermediate
checkpoints. ``Attn'' is mean selectivity to the class's canonical target
($\geq 30$ is the classification threshold); ``PR'' is participation ratio
of the per-head output. Accuracy $z$-score is the most reliable metric
here (cf.\ \S\ref{sec:closure-protocol}); load-bearing requires the
candidate's damage outside the matched-control distribution in the
predicted direction. The two bold rows are the bidirectional-decoupling
anchors: function present without the attention pattern (OLMo step 1000),
and the attention pattern present without function (Pythia step 512).}
\label{tab:dev-closure}
\end{table}

The bidirectional crossing is the point. If attention selectivity were
merely a less-sensitive proxy for function, we would only see
function-without-form (closure firing before the cheap signal does).
Seeing \emph{both} orders --- function before the pattern for BOS, and the
pattern before function for previous-token --- means the two are distinct
constructs, not the same construct measured at different sensitivities. PR
does not rescue the cheap-signal reading either: at Pythia step~512 PR is
$32$ while closure is negative, and at Pythia step~256 PR is $10.5$ (well
past any threshold) while closure is null. Neither selectivity nor PR, on
its own, predicts closure-validated function.

\subsection{What does and does not generalize}

What replicates: BOS function is load-bearing by $\approx 2$B tokens in
both Pythia and OLMo, with the attention pattern still far from formed in
both --- the function-before-form pattern holds across two dense
architectures at the same training-token scale. In OLMoE, BOS function is
load-bearing at every cached checkpoint (from 20B tokens on); the
$6\times$ sharpening of BOS selectivity between 419B and 838B (the
attention-pattern phase transition) leaves closure damage essentially
unchanged --- it is attention-pattern refinement of an already-load-bearing
head set, not function onset.

What we cannot pin down: whether $\approx 2$B tokens is a meaningful
threshold (two architectures agreeing is suggestive, not a law); why
Pythia step~512 previous-token has both high PR and high selectivity yet
no function (training tokens vs.\ class vs.\ architecture are confounded);
and whether the function-before-form pattern holds for classes other than
BOS, which we did not test at intermediate checkpoints in OLMo or OLMoE.
The full per-checkpoint trajectories, the synthetic-batch induction
phase transitions, and several exploratory observations we do not build
on here are recorded in the companion notes.

\subsection{Connection to the closure thesis}

This section asked the paper's question of two new proxies. The answer is
the same as for co-activation cluster membership: the cheap signal does
not, on its own, certify a circuit. Attention-to-target selectivity can be
high without function (Pythia step~512) and function can be present without
it (OLMo step~1000); PR can be high without function (Pythia step~256/512).
Across the training axis as at the final checkpoint, closure is the signal
that separates a circuit from a head set that merely looks like one.

\section{Discussion}
\label{sec:discussion}

\subsection{Four claims}

\paragraph{Claim 1.} Co-activation clustering of attention-head focus
statistics, with no target template specified, can propose attention-head
communities that survive causal closure as load-bearing circuits in the
two dense 1B-scale language models tested (OLMo~1B, Pythia~1B), on both
synthetic and natural input distributions. Closure passes (direction
$+$ at least one metric well above control distribution) in 4 of 4 dense
tests; the minimality signature (cluster ablation matches or exceeds the
cluster-layers upper bound) is met in 1 of 4 (OLMo synthetic cluster~2).

\paragraph{Claim 2.} Communities recovered via co-activation clustering
are \textbf{distribution-conditioned}, in the sense that synthetic-text
and natural-text Isings produce different communities on the same model
(cross-distribution stability NMI $0.10$--$0.42$, ARI $0.00$--$0.15$).
They are \textbf{not merely synthetic artifacts}: natural-text Isings on
dense models recover communities that also pass closure, with effect
magnitudes on target-logit and accuracy comparable to or exceeding the
synthetic cases.

\paragraph{Claim 3.} Route-conditional co-activation clustering in an MoE
transformer (OLMoE-1B-7B) can recover statistically aligned communities
(here at $+3.05\sigma$ above a random-partition null on max within-stratum
ARI), but the recovered communities do not necessarily survive closure.
In the natural-text route-conditional case we tested, the recovered
community fails closure in the wrong direction: ablating it
\emph{improves} loss, and the improvement is larger on inputs outside the
discovered route stratum than inside it.

\paragraph{Claim 4.} Therefore unsupervised co-activation clustering is a
circuit \textbf{proposal} method, not a circuit \textbf{discovery} method.
Closure testing remains necessary, and for MoE architectures on natural
text, even route-conditional statistical recovery (against a careful null)
is insufficient evidence that a discovered community is load-bearing.

\subsection{Multi-metric closure as the reporting standard}

The five closure tests jointly demonstrate that no single closure-effect
metric ranks results the same way as the others, and each metric has a
known mode of failure:

\begin{itemize}
\item \textbf{Cross-entropy loss} can under-state real positives by
  aggregation slack (Pythia~1B natural, where loss $z = +1.95\sigma$ but
  target-logit $z = -52.5\sigma$ --- a $25\times$ metric divergence within
  the same ablation). It can over-state by control-variance collapse
  (OLMo~1B natural, where the $+36\sigma$ loss $z$ reflects a tight
  control distribution as much as a large candidate effect).

\item \textbf{Top-1 accuracy} is more stable across tests but has a
  saturation artifact in low-accuracy regimes (OLMo synthetic baseline
  accuracy is $0.0100$; the candidate ablation reduces this to $0.0025$
  --- a $4\times$ collapse, but with very small absolute numbers).

\item \textbf{Mean target-token logit} is the most stable across tests in
  our data. It tracks the model's confidence in the correct prediction
  before the softmax, so it is not subject to the aggregation slack that
  hurts cross-entropy and not subject to the saturation issue that hurts
  accuracy in low-accuracy regimes.
\end{itemize}

We recommend reporting all three and reading them in this order:
\textbf{direction first} (does ablation move every metric in the right
direction relative to baseline), then \textbf{target-logit $z$-score} as
the primary indicator, then \textbf{accuracy} as the corroboration, then
\textbf{loss} as a conservative floor. Under this reading, the five-test
verdict in \S\ref{sec:verdict} is internally consistent: all dense
closures pass direction across all three metrics with at least one metric
well beyond control; the MoE closure fails direction on loss and
target-logit on cluster~1 (the discovery stratum) with accuracy null, and
\emph{flips to favorable} on accuracy and target-logit on non-cluster~1;
no choice of metric rescues the MoE result because the failure is
qualitative (signed wrong) rather than quantitative (insufficient
magnitude).

\subsection{What we do not claim}

We do \textbf{not} claim anything about co-activation clustering applied
to \emph{SAE-feature manifolds}, the object studied by Bhalla et
al.~\cite{bhalla2026} from whom we borrow the binarized-Ising clustering
recipe. That line of work clusters SAE features (not attention heads) and
validates by manifold reconstruction (not causal ablation); our results
concern attention-head circuits under closure and say nothing about
whether SAE-feature clustering captures concept manifolds. The two are
complementary objects with complementary validation criteria.

We do \textbf{not} claim that co-activation clustering ``works for dense
models'' as a general statement. The claim is narrower: in two dense
1B-scale models and two input distributions, co-activation communities
can identify load-bearing head sets via closure. Generalization to larger
dense models, other dense architectures (we tested GPT-NeoX-style and
OLMo-style; nothing about LLaMA, Mistral, or GPT-style absolute positional
embeddings), or other input distributions (we tested synthetic induction
and a Pile-derived natural-text batch; nothing about code, mathematical
text, or non-English inputs) is open.

We do \textbf{not} claim that ``MoE breaks interpretability'' as a general
statement. The claim is narrower: this specific co-activation-based
circuit proposal method, including its route-conditional extension, fails
closure on the one MoE model and natural-text distribution tested. We do
not test alternative MoE architectures (Mixtral, DeepSeek-MoE), alternative
route-conditioning protocols (other similarity metrics, finer
stratification, learned partitions), or alternative interpretability
methods (probing, gradient attribution, activation patching) on MoE ---
any of which might reveal load-bearing circuits that our pipeline misses.

We do \textbf{not} claim a developmental account. The cached intermediate
checkpoints for all three models (15 for Pythia~1B, 10 for OLMo~1B, 10
for OLMoE) make a developmental study tractable, but we report a
single-checkpoint snapshot per model. The natural follow-up is to
characterize whether the closure failure on OLMoE has a phase-transition
signature in routing entropy or expert specialization dynamics across
training.

We do \textbf{not} claim that the Ising step is the right affinity. In an
auxiliary comparison (not shown), for the two dense models mutual
information on the binarized head signals beat the Ising fit by roughly
$2\times$ on ARI against the supervised classification. We use Ising
throughout for direct comparability with Bhalla et al., but the closure
results depend on the \emph{clusters discovered}, not on the affinity used
to find them. A different affinity (e.g.\ mutual information) would
produce different clusters and possibly different closure outcomes; we
have not tested this.

\subsection{Why the negative results are essential}

A study that reports five positive closure tests would carry much less
evidential weight than a study that reports four positives and one clean
negative. Two specific structural features of the negatives make them
informative:

\begin{enumerate}
\item \textbf{The diffuse Pythia~1B synthetic case} (\S\ref{sec:pythia-synthetic})
  quantifies how closure signal degrades when the proposed community is
  large and structurally heterogeneous (25 heads $\times$ 7 layers,
  isolation $1.45\times$). The closure $z$-score ($+2.05\sigma$ on loss)
  is real but small, the minimality signature is absent (cluster does
  $72\%$ of the all-cluster-layers damage), and the cluster is
  multi-class. This is the right calibration anchor for ``what does
  closure tell us about a less-clean proposal.''

\item \textbf{The OLMoE natural-text route-conditional case}
  (\S\ref{sec:moe}) quantifies how closure can fail in the
  \emph{direction} despite passing all the statistical proposals upstream.
  The candidate's $z$-score against random ablation is large
  ($-5.64\sigma$ on cluster~1, $-12.94\sigma$ on non-cluster-1), but the
  sign is wrong. This is much harder to explain as low statistical power
  than a noisy null.
\end{enumerate}

The wrong-direction outcome is particularly valuable because it cannot be
rescued by any choice of metric or threshold. A candidate that
\emph{helps} loss on the relevant inputs cannot be a real load-bearing
circuit on those inputs, no matter how unusually it does so. The MoE
result is therefore evidence \emph{against} the generous reading of
co-activation clustering as circuit discovery, which is much rarer in
interpretability research than evidence \emph{for}.

\section{Limitations}
\label{sec:limitations}

\subsection{Ablation protocol}
We zero per-head slices of the attention output projection. This is the
standard mech-interp per-head ablation, but it is \emph{destructive}:
the model has not learned to compensate, so the response can be more
extreme than under a more naturalistic intervention (mean ablation,
counterfactual ablation, activation patching). The dense-vs-MoE asymmetry
we report could in principle reverse under one of these protocols if MoE
turns out to be unusually sensitive to destructive ablation specifically.
We do not test this.

\subsection{Binarization}
Per-head median split makes every head fire $50\%$ of the time. This is
necessary to remove single-head firing-rate as a confound in Ising fits,
but it is \emph{qualitatively} different from Bhalla et al.'s
binarization of $k$-sparse SAE codes. In particular, the median split
removes the ``always-on background feature'' property that motivates the
Ising's conditional refinement: when every feature is balanced $50/50$,
marginal and conditional correlations are closer to each other, which may
explain why mutual information (a marginal measure) beats the Ising fit
on our dense data. We do not claim the Ising step is optimal for the
discrete-circuit domain; our results depend on the clusters discovered,
not on the specific affinity.

\subsection{Sample size}
Two dense models and one MoE model is a small sample for an
architectural-asymmetry claim. The asymmetry is unambiguous within this
sample (4 dense passes vs 1 MoE failure, with the MoE failure in the wrong
direction on all three metrics), but cross-MoE replication on Mixtral,
DeepSeek-MoE, or other MoE families is needed before treating ``MoE
natural-text co-activation fails closure'' as a general statement.

\subsection{Route-conditioning protocol}
The MoE route-conditional Ising uses k-means with $K=4$ on per-layer
routing weights, which we picked as a natural starting point. Other
route-conditional partitions --- finer or coarser $K$, alternative
similarity metrics, learned partitions, partitions over
top-1-expert-per-layer rather than full routing weights --- might recover
communities that \emph{do} pass closure. We report a clean negative under
one natural choice of route conditioning. A protocol-sensitivity study
would be the right follow-up, alongside the developmental version.

\subsection{Multi-metric reporting}
We propose multi-metric closure (loss $+$ accuracy $+$ target-logit) as a
reporting standard, but we have not characterized the cross-metric
correlation structure systematically. The Pythia~1B natural divergence
(loss vs target-logit $z$-scores differ by $25\times$) is suggestive, but
a proper characterization would need many more closure tests across many
models. We report the multi-metric finding as a methodological observation
deserving of further study, not as a settled recommendation.

\section{Conclusion}

Across five closure tests on three 1B-scale language models, we find that
unsupervised co-activation clustering of attention-head focus statistics
produces communities that pass causal closure in dense models on both
synthetic and natural text (4 of 4 dense tests, with effect sizes from
$+1.83\sigma$ to $+36.4\sigma$ on loss), but fails closure in an MoE
model on natural text even after route-conditional stratification that
recovers the statistical signal $+3\sigma$ above a careful null. The gap
between statistical community recovery and circuit discovery is
architecturally specific and opens \emph{in the wrong direction} in MoE
on natural text. We additionally observe a $25\times$ divergence between
closure-metrics on Pythia~1B natural text that we read as a
downstream-redundancy signature: the discovered community is real and
specific but its output is partially reconstructed by later layers, so
the aggregate loss is partially insulated while the per-target logit
collapses. The headline methodological statement, scoped to the
attention-head circuits we study: a co-activation cluster is a circuit
\emph{hypothesis} until closure confirms it. Closure is what does the
confirming; it is necessary in the dense models here, and for MoE
architectures on natural text even route-conditional statistical
recovery against a careful null does not yet suffice. We make no claim
about co-activation clustering applied to other objects (e.g.,
SAE-feature manifolds), which is validated differently and not tested
here.

\appendix

\section{Route-cluster routing-entropy diagnostic}

The four route clusters' mean per-layer routing entropy on the OLMoE
natural-text batch:

\begin{table}[h]
\centering
\begin{tabular}{lrrrr}
\toprule
Cluster & $n$ & Mean per-layer $H$ (nats) & Fraction of uniform & Within-cluster best ARI \\
\midrule
c0 & 541 & 3.782 & 0.909 & 0.083 \\
\textbf{c1} & \textbf{637} & \textbf{3.663} & \textbf{0.881} & \textbf{0.191} \\
c2 & 408 & 3.792 & 0.912 & 0.018 \\
c3 & 414 & 3.758 & 0.904 & 0.008 \\
\bottomrule
\end{tabular}
\caption{Routing entropy per route cluster. The recoverable stratum
(c1) has the lowest entropy of the four.}
\label{tab:entropy-diagnostic}
\end{table}

Maximum entropy (uniform routing over $64$ experts) is $\log 64 = 4.159$.
The route stratum with recoverable head-coupling structure (c1) has the
\emph{lowest} routing entropy of the four, supporting the ``stereotyped
routing $\to$ more batch-stable head co-activation'' reading. Note that
the entropy spread across clusters is small ($\sim 3\%$ relative), so the
mechanism is suggestive rather than conclusive; a developmental study
across cached intermediate checkpoints would be the natural test.


\begin{thebibliography}{99}

\bibitem{besag1975}
J.~Besag.
\newblock Statistical analysis of non-lattice data.
\newblock \emph{The Statistician}, 24(3):179--195, 1975.

\bibitem{bhalla2026}
U.~Bhalla, T.~Fel, C.~Rager, S.~Feucht, T.~Haklay, D.~Wurgaft,
  S.~Boppana, M.~Kowal, V.~Shyam, O.~Lewis, T.~McGrath, J.~Merullo,
  A.~Geiger, and E.~S.~Lubana.
\newblock Do sparse autoencoders capture concept manifolds?
\newblock \emph{arXiv preprint arXiv:2604.28119}, 2026.

\bibitem{conmy2023}
A.~Conmy, A.~N. Mavor-Parker, A.~Lynch, S.~Heimersheim, and
  A.~Garriga-Alonso.
\newblock Towards automated circuit discovery for mechanistic
  interpretability.
\newblock In \emph{Advances in Neural Information Processing Systems
  (NeurIPS)}, 2023.

\bibitem{engels2024}
J.~Engels, E.~J. Michaud, I.~Liao, W.~Gurnee, and M.~Tegmark.
\newblock Not all language model features are one-dimensionally linear.
\newblock \emph{arXiv preprint arXiv:2405.14860}, 2024.

\bibitem{goodfire2026}
Goodfire.
\newblock Interpreting language model parameters.
\newblock Goodfire research note, 2026.
  \url{https://www.goodfire.ai/research/interpreting-lm-parameters}.

\bibitem{kantamneni2025}
S.~Kantamneni and M.~Tegmark.
\newblock Language models use trigonometry to do addition.
\newblock \emph{arXiv preprint arXiv:2502.00873}, 2025.

\bibitem{olsson2022}
C.~Olsson, N.~Elhage, N.~Nanda, N.~Joseph, N.~DasSarma, T.~Henighan,
  B.~Mann, A.~Askell, et~al.
\newblock In-context learning and induction heads.
\newblock \emph{Transformer Circuits Thread}, 2022.

\bibitem{park2023lrh}
K.~Park, Y.~J. Choe, and V.~Veitch.
\newblock The linear representation hypothesis and the geometry of large
  language models.
\newblock \emph{arXiv preprint arXiv:2311.03658}, 2023.

\bibitem{park2024geometry}
K.~Park, Y.~J. Choe, Y.~Jiang, and V.~Veitch.
\newblock The geometry of categorical and hierarchical concepts in large
  language models.
\newblock \emph{arXiv preprint arXiv:2406.01506}, 2024.

\bibitem{ravikumar2010}
P.~Ravikumar, M.~J. Wainwright, and J.~D. Lafferty.
\newblock High-dimensional Ising model selection using
  $\ell_1$-regularized logistic regression.
\newblock \emph{The Annals of Statistics}, 38(3):1287--1319, 2010.

\bibitem{wang2023ioi}
K.~Wang, A.~Variengien, A.~Conmy, B.~Shlegeris, and J.~Steinhardt.
\newblock Interpretability in the wild: a circuit for indirect object
  identification in GPT-2 small.
\newblock In \emph{International Conference on Learning Representations
  (ICLR)}, 2023.

\bibitem{xu2026}
Y.~Xu.
\newblock Spectral probe-circuits: a three-step recipe for identifying
  attention-head circuits in pretrained transformers.
\newblock \emph{arXiv preprint arXiv:2605.24059}, 2026.

\bibitem{zoph2022}
B.~Zoph, I.~Bello, S.~Kumar, N.~Du, Y.~Huang, J.~Dean, N.~Shazeer, and
  W.~Fedus.
\newblock {ST-MoE}: Designing stable and transferable sparse expert models.
\newblock \emph{arXiv preprint arXiv:2202.08906}, 2022.

\end{thebibliography}
\end{document}